\documentclass{article}
\usepackage[preprint]{neurips_2025}
\usepackage[T1]{fontenc}
\usepackage{microtype}
\usepackage{amsmath,amssymb,booktabs,tabularx,array}
\usepackage[table]{xcolor}
\usepackage{graphicx,caption,subcaption,enumitem}
\usepackage{tikz,pgfplots}
\usetikzlibrary{arrows.meta,positioning,calc}
\usepgfplotslibrary{groupplots}
\pgfplotsset{compat=1.18}
\usepackage[hidelinks]{hyperref}
\definecolor{TPDOrange}{HTML}{C77525}
\definecolor{ActionBlue}{HTML}{277DA8}
\definecolor{ReasonGray}{HTML}{707780}
\definecolor{ConstrainedPurple}{HTML}{8064A2}
\definecolor{ConstantGreen}{HTML}{458A74}
\definecolor{ExpBlue}{HTML}{326A9F}
\definecolor{ExpOrange}{HTML}{C76B24}
\definecolor{ExpGray}{HTML}{68717D}
\definecolor{ExpPaleBlue}{HTML}{EAF2FA}
\definecolor{ExpPaleOrange}{HTML}{FFF3E5}
\newcommand{\tpd}{\textsc{TPD}}
\newcommand{\acon}{\ensuremath{A_{\mathrm{constr}}}}
\newcommand{\bconst}{\ensuremath{B_{\mathrm{const}}}}
\newcommand{\bshuffle}{\ensuremath{B_{\mathrm{shuffle}}}}

\newcommand{\nr}{\textsc{nr}}
\newcommand{\uv}{\textsc{uv}}
\newcolumntype{Y}{>{\raggedright\arraybackslash}X}
\setlist[itemize]{leftmargin=*,itemsep=2pt,topsep=4pt}
\pgfplotsset{budgetaxis/.style={
 width=0.47\textwidth,height=5.7cm,xmode=log,log basis x=2,
 xmin=80,xmax=1000,ymin=-3,ymax=100,
 xtick={98,200,404,808},xticklabels={98,200,404,808},
 ytick={0,20,40,60,80,100},xlabel={Teacher demonstrations},
 grid=major,grid style={black!8},axis line style={black!45},
 tick label style={font=\small},label style={font=\small},
 title style={font=\small\bfseries},legend style={font=\scriptsize,draw=none,fill=white},
 legend pos=north west,mark size=2.2pt}}
\title{Learning to Act with Task Progress:\\
    \large Distilling Small Agents from Compact Teacher Supervision}
\author{
    Wenxi Gan \\
    City University of Hong Kong \\
    Shenzhen Loop Area Institute \\
    \texttt{wenxigan2-c@my.cityu.edu.hk}
}
\date{}
\begin{document}
\maketitle

\begin{abstract}
Learning from large-model demonstrations offers a way to train
small agents that can complete recurring tasks without calling
a large model at every step.
A central design choice is what to retain from teacher trajectories
that contain reasoning, actions, and information about task progress.
We introduce Task-Progress Distillation (TPD), an offline approach
that pairs each demonstrated action with a short label describing
the current task stage.
The student learns these compact targets and selects actions by
jointly scoring admissible stage--action pairs, which a deterministic
harness executes in the environment.
On ALFWorld, a 1.7B student trained with 404 demonstrations achieves
72.4\% mean unseen task success with either TPD or action-only
supervision, compared with 48.3\% for a reasoning-trained student
using constrained action selection.
Explicit stages provide an additional benefit at 200 demonstrations,
improving success from 48.0\% to 67.7\% over action-only supervision.
With more demonstrations, the action-only student closes the gap,
and both approaches reach 76.9\% at 808 demonstrations.
Shared-history analyses link part of TPD's local advantage to
better decisions when moving between subgoals, particularly from
object acquisition to processing.
These results show that compact supervision can train effective
small task agents, while explicit task progress provides additional
guidance at an intermediate demonstration budget.
\end{abstract}

\section{Introduction}
\label{sec:introduction}
For recurring tasks, a large model can provide demonstrations from which a smaller model learns to act on its own.
The aim is for the student to complete new instances without asking the teacher to choose each action.
The teacher trajectories in our setting contain reasoning, actions, and feedback that records how the task progresses.
With limited demonstrations, the choice of what the student should learn becomes central.

Action imitation already provides a useful foundation.
LM-BUTLER shows that action modeling can train effective ALFWorld agents, while FireAct and AgentTuning learn from language-model trajectories and broader agent data \cite{lmbutler,fireact,agenttuning}.
Other methods separate reasoning and action losses or adjust token updates to change how the student learns from each trajectory \cite{sad,opcsft}.
Online learning offers another source of improvement: the student can collect experience beyond the initial demonstrations \cite{gigpo,hiper}.
We keep the student's demonstrations fixed and compare which training targets best support task execution.

Our approach combines compact supervision with constrained action selection.
We train the student to predict the next executable action from the task instruction and action--feedback history.
At execution time, the student scores the actions allowed by the environment and selects one to execute.
The action-only variant (C) uses this action as its entire decision target.
Task-Progress Distillation (TPD, arm B) adds a short label for the current task stage: finding the target, processing it, or delivering it.
TPD scores stage--action pairs; a deterministic execution harness sends the selected action to the environment and records its feedback.

Stage labels address a common difficulty in multi-step tasks: moving from one subgoal to the next.
Suppose the goal is to heat a cup and place it on a table.
Once the cup has been acquired, the next subgoal is to take it to the microwave.
An agent can instead keep searching, choosing actions that are allowed but no longer serve the immediate subgoal.
Labeling this decision as processing tells the student which subgoal the next action should serve.

Both compact variants learn to complete tasks from a few hundred demonstrations.
On ALFWorld, a Qwen3-1.7B student reaches 72.4\% mean unseen success from 404 demonstrations with either action-only supervision or TPD; both reach 76.9\% with 808 demonstrations.
A student trained on reasoning and actions reaches 5.0\% at the 404-demonstration budget with free generation, and 48.3\% when the same checkpoints use constrained action selection.
Constrained selection recovers much of the reasoning-trained student's performance, but both compact variants remain ahead under their respective training and execution configurations.

The additional benefit of task stages depends on the demonstration budget.
With 200 demonstrations, TPD reaches 67.7\% unseen success compared with 48.0\% for action-only supervision, a gain of 19.7 percentage points.
The action-only student catches up in mean unseen success at 404 and 808 demonstrations.
We therefore examine both when stages help and which decisions change when the student uses them.

We investigate this question by comparing trajectories and then testing decisions under the same interaction history.
In processing tasks, one recurring difference appears after the correct object has been acquired: TPD more often starts the required operation.
On selected seen histories, fixing a stage that matches the task's progress more often leads to an action that advances the task than fixing a mismatched stage.
This sensitivity overlaps with some decisions where TPD makes progress and C does not, mainly in a few heating tasks and with results that depend on the scoring rule.
Across three training seeds on seen tasks, shuffling stage labels across histories also lowers success relative to aligned labels.
Together, the results suggest that task-progress labels help the student learn some transitions between subgoals; with more action demonstrations, C increasingly makes similar choices.

Our contributions are:
\begin{itemize}
\item A compact distillation pipeline that combines action-focused supervision, task-stage labels, and admissible-action scoring to train a small agent from fixed teacher demonstrations.
\item A comparison of reasoning, action-only, and task-progress configurations across demonstration budgets, showing strong compact-agent performance and an additional stage benefit at 200 demonstrations.
\item A behavioral analysis linking part of TPD's local advantage to stage-sensitive task transitions, with smaller B/C decision differences as demonstrations increase.
\end{itemize}

\section{Related Work}
\label{sec:related}
\paragraph{Learning agents from demonstrations.}
LM-BUTLER trains language models to predict expert actions and reports strong ALFWorld results, including a few-demonstration setting \cite{lmbutler}.
FireAct distills trajectories produced by language-model agents, and AgentTuning combines agent interactions with general instruction data \cite{fireact,agenttuning}.
Our action-only student follows this line of work.
It provides a practical agent in its own right and a baseline for measuring the added value of task-stage labels.

These methods also differ in what they ask the student to reproduce.
Structured Agent Distillation separates reasoning and action segments, while OPC-SFT changes the updates assigned to low-probability tokens \cite{sad,opcsft}.
TPD retains the demonstrated action and replaces the long reasoning target with a short task-stage label.
Our reasoning baseline uses ordinary completion supervision; SAD and OPC-SFT introduce specialized training objectives \cite{sad,opcsft}.
SAD also describes online teacher queries during training \cite{sad}; our student learns from a fixed dataset without further teacher queries.

\paragraph{Learning through additional interaction.}
A fixed dataset may omit states that a student reaches after making a mistake.
Several methods address this by collecting more experience during training.
ETO adds student exploration after SFT, A$^3$T combines reasoning annotation with self-training, and HSL relabels collected trajectories \cite{eto,a3t,hsl}.
HPL constructs preferences over trajectories, steps, and action groups, while SAGE-OPD provides selective teacher assistance during on-policy distillation \cite{hpl,sageopd}.
These methods let the training data reflect the student's own behavior.

Online agent RL uses rewards from interaction to update the policy.
GiGPO groups decisions to estimate local learning signals, HiPER assigns credit at different planning levels, and STAPO selectively optimizes steps according to their dependence on the trajectory \cite{gigpo,hiper,stapo}.
RLVMR uses verifiable meta-reasoning rewards, and EnvRL adds objectives based on environment dynamics \cite{rlvmr,envrl}.
Our study instead asks what the student can learn from a fixed set of demonstrations, without collecting additional training rollouts.
We use the environment to collect teacher demonstrations and evaluate the student, while student weight updates use only the fixed dataset.

\paragraph{Task structure and progress.}
Task structure can guide learning as well as execution.
HiPER separates planning from execution, and BEACON uses milestones to construct rewards and advantages \cite{hiper,beacon}.
TPD uses three task-progress labels as supervised targets and as context when scoring the next action.
From History to State studies compact state-based context, while Automata from Agent Traces extracts finite-state structure for prediction and monitoring \cite{historytostate,automata}.
Our student retains the full action--feedback history.
The stage identifies the current subgoal; the full history still supplies the objects, locations, and feedback needed to choose an action.

\section{Method}
\label{sec:method}

\subsection{Overview and Task Interface}
\begin{figure}[t]
\centering
\includegraphics[width=\linewidth]{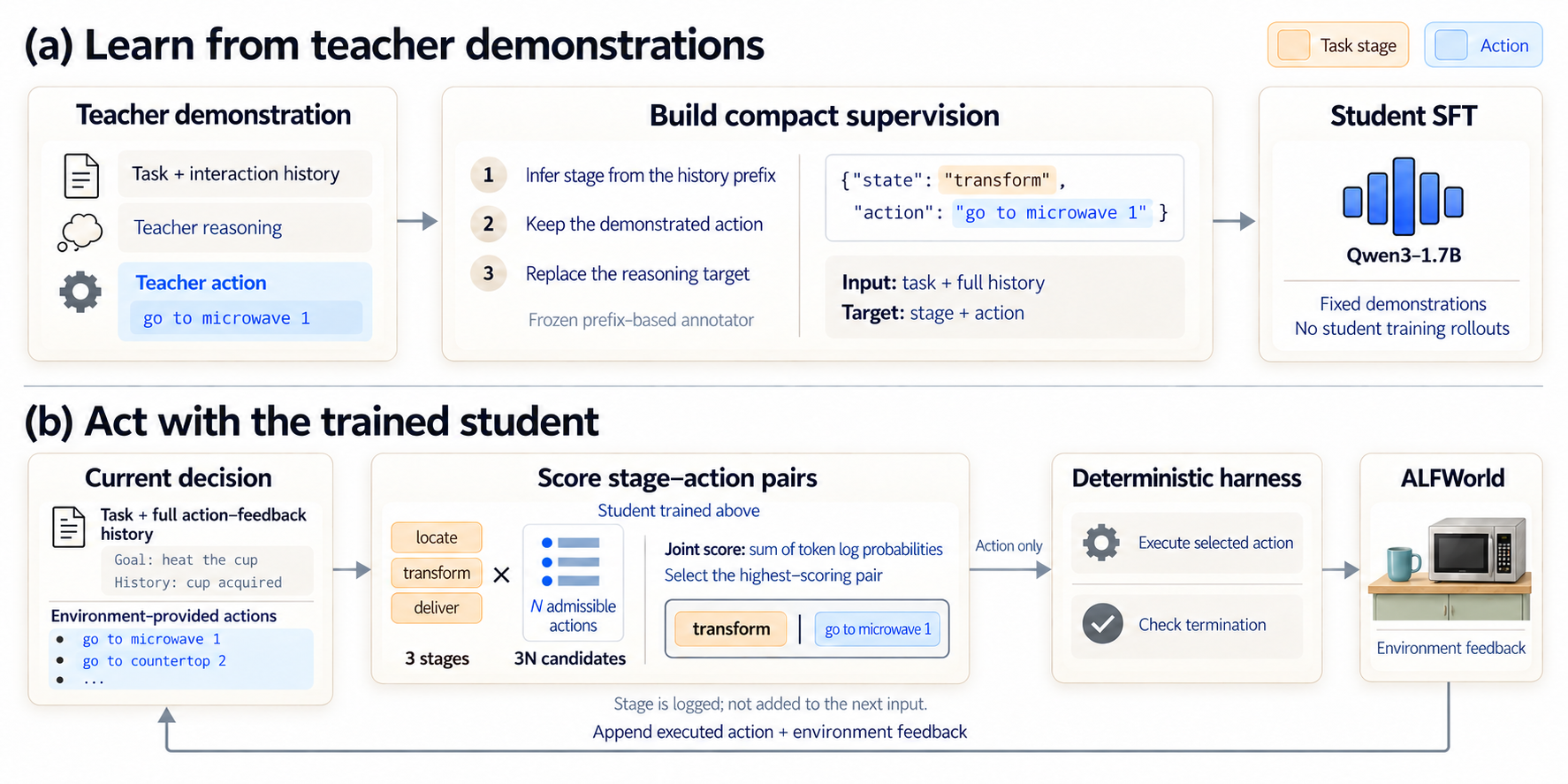}
\caption{\textbf{Offline training and closed-loop execution.}
(a) A frozen annotator reads each teacher-history prefix and pairs the current task stage with the demonstrated action for student SFT.
(b) The student scores $3|\mathcal A(h_t)|$ stage--action candidates, and the harness executes the action from the highest-scoring pair.
The next input includes the action and environment feedback; the predicted stage is logged only for analysis.}

\label{fig:overview}
\end{figure}

The pipeline has an offline training stage and a closed-loop execution stage (Figure~\ref{fig:overview}).
Offline, we convert teacher demonstrations into compact stage--action targets and fine-tune the student.
During execution, the student scores admissible stage--action pairs. The harness sends the selected action to the environment and appends the action and feedback to the history.

Let $g$ be the goal and
\begin{equation}
 h_t=(g,o_0,a_0,o_1,\ldots,a_{t-1},o_t)
\end{equation}
the history available before action $a_t$.
The environment provides the admissible action set $\mathcal A(h_t)$.
We form the model input $x_t$ from the history and an instruction specifying the required output format.
B and C use the same training histories and demonstrated actions; B additionally predicts a task-stage label.

\subsection{Compact Decision Targets}
The simplest target is the action selected by the teacher.
The action-only student learns this target from the goal and history.
TPD also predicts the current task stage.
For each teacher decision, a frozen annotator assigns
\begin{equation}
 s_t=f(g,h_t),\qquad
 s_t\in\mathcal S=\{\texttt{locate},\texttt{transform},\texttt{deliver}\},
\end{equation}
where the stages distinguish finding the target, carrying out required processing, and completing delivery.
For example, after the agent acquires a cup that still needs heating, the label is \texttt{transform}.

The annotator tracks whether the target is held, whether processing has succeeded, and whether placement has occurred.
It updates these flags from successful environment feedback and combines them with the task type.
It labels each decision using only feedback available before that action, so the label describes progress at decision time.
Appendix~\ref{app:implementation-details} gives the implementation details.

A TPD training target can therefore be as short as
\begin{center}
\small\texttt{\{"state":"transform","action":"go to microwave 1"\}}.
\end{center}
The action-only target contains the same action without the stage field.
Both targets specify the next action; TPD also identifies its task stage.

We train each model with completion-only supervised learning.
Given a target sequence $y_t$, the objective minimizes the negative log probability of its supervised tokens:
\begin{equation}
 \mathcal L(\theta)\ \propto\
 -\sum_{(x_t,y_t)\in\mathcal D}\sum_{j\in\mathcal M_t}
 \log p_\theta(y_{t,j}\mid x_t,y_{t,<j}),
 \label{eq:sft}
\end{equation}
where $\mathcal M_t$ contains completion positions and excludes system and user tokens.
We apply token-level cross-entropy to the supervised completion tokens.
For TPD, the entire completion---stage, action, and template tokens---is trained jointly, with no separate segment weights.
Longer reasoning targets devote a smaller fraction of supervised tokens to the action.
Training and candidate scoring use the same prompt templates and compact serialization.

\subsection{Joint Candidate Scoring and the Harness}
At execution time, TPD chooses a stage and an action together.
For each admissible action, we construct one candidate completion for each of the three stages.
Let $y(s,a)$ be the completion for stage $s$ and action $a$.
The student scores the pair by summing its completion-token log probabilities:
\begin{equation}
 q_\theta(s,a\mid h_t)=
 \sum_{j=1}^{|y(s,a)|}\log p_\theta(y_j(s,a)\mid x_t,y_{<j}(s,a)).
 \label{eq:jointscore}
\end{equation}
It then selects the highest-scoring pair:
\begin{equation}
 (\hat s_t,\hat a_t)=
 \mathop{\arg\max}_{s\in\mathcal S,\,a\in\mathcal A(h_t)}
 q_\theta(s,a\mid h_t).
 \label{eq:selection}
\end{equation}
With $N$ admissible actions, TPD scores $3N$ candidates.
The action-only student scores $N$, using the same sum of log probabilities without length normalization.
Candidate order and tie-breaking are fixed (Appendix~\ref{app:implementation-details}).

The harness executes only $\hat a_t$ and appends that action and its feedback to the history.
The next decision is made from this updated history, with no teacher call.
Predicted stages are logged for analysis but are not carried into the next input.
Within a decision, the stage affects the context used to score action tokens and contributes to the joint score. Across decisions, the model receives only the action--feedback history.
Execution stops when the environment terminates the episode or after 50 steps.

\section{Experiments}
\label{sec:experiments}
\subsection{Experimental Setup}
\label{sec:setup}
\paragraph{Environment and demonstrations.}
We evaluate multi-step task execution in ALFWorld \cite{alfworld}.
The six task families cover picking and placing an object, picking two objects, examining an object under light, and cleaning, heating, or cooling an object.
We train Qwen3-1.7B \cite{qwen3} on successful demonstrations collected from DeepSeek-v4-pro with the \texttt{teacher\_v2} prompt and temperature 0.7.
The four datasets contain 98, 200, 404, and 808 trajectories.
The 200-, 404-, and 808-trajectory datasets contain 3,712, 7,007, and 13,618 decision records, respectively.
The 808-trajectory dataset retains all original 404 demonstrations and adds 404 more using the planned allocation across task categories.

\paragraph{Training and model selection.}
Each independently trained model starts from the same Qwen3-1.7B checkpoint and undergoes three epochs of full-parameter SFT.
We use a learning rate of $10^{-5}$, an effective batch size of 32, and a maximum sequence length of 8,192.
Within each budget, the three seeds (42, 1234, and 2026) share one dataset, and the arms use matching histories and demonstrated actions.
After each epoch, we measure action agreement on a seen probe containing 697 decisions from 35 tasks.
We select the checkpoint with the highest probe action agreement, breaking ties in favor of the earlier epoch.
Appendix~\ref{app:implementation-details} provides the full recipe and software versions.

\paragraph{Comparisons.}
Table~\ref{tab:arms} separates the training targets from the execution interfaces.
A is trained to reproduce the teacher's reasoning and action, and freely generates up to 500 tokens per decision.
If parsing fails, the raw output is sent to the environment and still consumes a step.
\acon{} uses the same A checkpoint, generates reasoning, and selects an action by scoring admissible candidates.
This comparison measures the effect of changing execution with the weights held fixed.
C is trained on action-only targets; B uses TPD's stage--action targets and joint scoring.
The constant-stage and shuffled-stage controls test what the stage labels contribute.

\begin{table}[t]
\centering\small
\caption{Training targets and execution interfaces. $N$ denotes the current admissible action count. A and \acon{} share weights at the corresponding budget and seed; other trained arms are initialized independently from the same
Qwen3-1.7B checkpoint.}
\label{tab:arms}
\begin{tabularx}{\textwidth}{@{}lYY@{}}
\toprule
Arm & Supervised target & Execution interface \\
\midrule
A & Teacher reasoning and action & Free generation followed by action parsing \\
\acon & Same checkpoint as A & Generate reasoning, then score $N$ admissible actions \\
B / \tpd & Prefix-aligned stage and action & Joint scoring over $3N$ stage--action pairs \\
C & Action only & Score $N$ action candidates \\
\bconst & Constant \texttt{phase} field and action & Score $N$ fixed-stage action candidates \\
\bshuffle & Shuffled stage labels and unchanged actions & Same $3N$ candidate structure as B \\
\bottomrule
\end{tabularx}
\end{table}

\paragraph{Evaluation sets and metrics.}
Our main evaluation covers all 134 official \texttt{valid\_unseen} tasks, with a maximum of 50 actions per episode.
For validation and diagnosis, we use the 116 of 140 official \texttt{valid\_seen} tasks that a rule-based expert completed within 50 steps.
The 35-task checkpoint-selection probe is drawn from these 116 tasks.
We therefore use seen results for validation and diagnosis, and the full unseen split for the main performance comparison.

We report task success, steps per episode, and recorded output tokens.
Results are means and sample standard deviations over three training seeds unless otherwise stated; each seed is evaluated on the same tasks.
Free-generation student evaluation uses temperature zero.
We also report open-loop action agreement with a rule-based expert reference.
Open-loop agreement compares a model's next action with the expert action on a supplied history; a different action may still advance the task.

\paragraph{Experiment scope.}
The planned main comparison comprises the original A/B/C matrix and the 404-demonstration evaluation of \acon{}.
An early inspection of unseen aggregates was documented without changing the original run plan; checkpoint selection used the seen probe.
The later data extension, additional \acon{} evaluations, stage controls, and prefix analyses are exploratory.
The shuffled-stage control evaluates all three training seeds on seen tasks.

\subsection{Compact Supervision Trains Effective Small Agents}
\label{sec:overall}
\paragraph{Small agents trained from a few hundred demonstrations.}
We first measure how often the compact students complete unseen tasks.
With 404 demonstrations, both C and TPD reach 72.4\% mean unseen success; with 808, both reach 76.9\% (Table~\ref{tab:main}).
Both variants train on fixed teacher demonstrations and act without teacher assistance during evaluation.
The teacher reference reaches 88.1\% (118/134), leaving an 11.2-point gap at 808 demonstrations.
The strong action-only result shows that compact action supervision already provides a useful agent before adding task-stage labels.

\begin{table}[t]
\centering
\small
\setlength{\tabcolsep}{7pt}
\renewcommand{\arraystretch}{1.16}
\caption{Unseen success (\%), mean $\pm$ sample standard deviation across three training seeds, each evaluated on 134 tasks.
Blue and orange identify C and TPD; bold marks the highest observed mean in each column, including ties.}

\label{tab:main}
\begin{tabular}{@{\hspace{\tabcolsep}}lcccc@{\hspace{\tabcolsep}}}
\toprule
& \multicolumn{4}{c}{Number of training demonstrations} \\
\cmidrule(lr){2-5}
Configuration & 98 & 200 & 404 & 808 \\
\midrule
A: reasoning--action
& $0.0\pm0.0$
& $0.0\pm0.0$
& $5.0\pm1.7$
& $16.4\pm1.5$ \\
\acon
& $1.5\pm0.8$
& $30.8\pm6.5$
& $48.3\pm4.8$
& $48.8\pm5.6$ \\
\rowcolor{ExpPaleBlue}
C: action only
& $42.5\pm2.0$
& $48.0\pm12.1$
& $72.4\pm1.5$
& $76.9\pm3.0$ \\
\bconst
& $31.6\pm4.4$
& $51.5\pm11.7$
& $\mathbf{72.6}\pm1.6$
& $\mathbf{77.1}\pm6.8$ \\
\rowcolor{ExpPaleOrange}
B: stage--action (\tpd)
& $\mathbf{48.0}\pm2.8$
& $\mathbf{67.7}\pm2.4$
& $72.4\pm4.5$
& $76.9\pm3.3$ \\
\midrule
B $-$ C (pp)
& $+5.5$ & $+19.7$ & $0.0$ & $0.0$ \\
B $-$ \bconst{} (pp)
& $+16.4$ & $+16.2$ & $-0.2$ & $-0.2$ \\
\midrule
Teacher reference
& \multicolumn{4}{c}{$88.1$ (118/134; one teacher evaluation configuration)} \\
\bottomrule
\end{tabular}
\end{table}

\begin{figure}[t]
\centering
\begin{tikzpicture}
\begin{groupplot}[
    group style={group size=2 by 1,horizontal sep=1.2cm},
    width=0.485\textwidth,
    height=6.4cm,
    xmode=log,
    log basis x=2,
    xmin=80,
    xmax=1000,
    ymin=-3,
    ymax=100,
    xlabel={Teacher demonstrations},
    xtick={98,200,404,808},
    xticklabels={98,200,404,808},
    ytick={0,20,40,60,80,100},
    tick label style={font=\small},
    label style={font=\small},
    title style={font=\small\bfseries},
    axis line style={draw=black!55},
    tick style={draw=black!55},
    axis x line*=bottom,
    axis y line*=left,
    grid=major,
    grid style={draw=black!9},
    legend style={
        font=\scriptsize,
        draw=none,
        fill=white,
        fill opacity=0.93,
        text opacity=1,
        cells={anchor=west}
    },
    legend pos=north west
]

\nextgroupplot[title={(a) Supervision and execution},ylabel={Unseen task success (\%)}]

\addplot+[ExpGray,thick,mark=square*,mark options={fill=ExpGray,draw=ExpGray},
    error bars/.cd,y dir=both,y explicit,
    error bar style={line width=0.8pt},
    error mark options={rotate=90,mark size=3pt}]
    table[x=x,y=y,y error=e] {
x y e
98 0 0
200 0 0
404 4.975 1.723
808 16.4 1.5
};
\addlegendentry{A}

\addplot+[ConstrainedPurple,thick,mark=diamond*,mark options={fill=ConstrainedPurple,draw=ConstrainedPurple},
    error bars/.cd,y dir=both,y explicit,
    error bar style={line width=0.8pt},
    error mark options={rotate=90,mark size=3pt}]
    table[x=x,y=y,y error=e] {
x y e
98 1.467 0.751
200 30.8 6.5
404 48.259 4.798
808 48.8 5.6
};
\addlegendentry{\acon}

\addplot+[ExpBlue,thick,mark=*,mark size=2.7pt,mark options={fill=ExpBlue,draw=ExpBlue},
    error bars/.cd,y dir=both,y explicit,
    error bar style={line width=0.8pt},
    error mark options={rotate=90,mark size=3pt}]
    table[x=x,y=y,y error=e] {
x y e
98 42.537 1.974
200 48.010 12.064
404 72.388 1.493
808 76.866 2.985
};
\addlegendentry{C}

\draw[<->,>=Stealth,draw=ExpGray,line width=0.8pt]
    (axis cs:450,48.259) -- (axis cs:450,72.388);
\node[anchor=west,font=\scriptsize,text=ExpGray,fill=white,inner sep=2pt]
    at (axis cs:458,60) {$+24.1$ pp};

\nextgroupplot[title={(b) The role of task progress}]

\addplot+[ExpBlue,thick,mark=*,mark size=2.7pt,mark options={fill=ExpBlue,draw=ExpBlue},
    error bars/.cd,y dir=both,y explicit,
    error bar style={line width=0.8pt},
    error mark options={rotate=90,mark size=3pt}]
    table[x=x,y=y,y error=e] {
x y e
98 42.537 1.974
200 48.010 12.064
404 72.388 1.493
808 76.866 2.985
};
\addlegendentry{C}

\addplot+[ExpOrange,thick,mark=square,mark size=4.0pt,
    mark options={solid,fill=none,draw=ExpOrange,line width=1.1pt},
    error bars/.cd,y dir=both,y explicit,
    error bar style={line width=0.8pt},
    error mark options={rotate=90,mark size=4pt}]
    table[x=x,y=y,y error=e] {
x y e
98 48.010 2.825
200 67.662 2.399
404 72.388 4.539
808 76.866 3.253
};
\addlegendentry{B}

\addplot+[ConstantGreen,thick,dashed,mark=square*,mark options={fill=ConstantGreen,draw=ConstantGreen},
    error bars/.cd,y dir=both,y explicit,
    error bar style={line width=0.8pt},
    error mark options={rotate=90,mark size=3pt}]
    table[x=x,y=y,y error=e] {
x y e
98 31.592 4.373
200 51.493 11.657
404 72.6 1.6
808 77.1 6.8
};
\addlegendentry{B\textsubscript{const}}

\addplot[only marks,mark=square,mark size=1.4pt,draw=ExpOrange,fill=none,opacity=.5,forget plot]
    coordinates {(98,44.776)(98,50)(98,49.254)(200,64.925)(200,69.403)(200,68.657)(404,69.403)(404,70.149)(404,77.612)(808,73.134)(808,78.358)(808,79.104)};
\addplot[only marks,mark=*,mark size=1pt,ExpBlue,opacity=.45,forget plot]
    coordinates {(98,40.299)(98,44.030)(98,43.284)(200,35.075)(200,50)(200,58.955)(404,73.881)(404,70.896)(404,72.388)(808,76.866)(808,79.851)(808,73.881)};

\draw[<->,>=Stealth,draw=ExpOrange,line width=0.8pt]
    (axis cs:230,48.010) -- (axis cs:230,67.662);
\node[anchor=west,font=\scriptsize,text=ExpOrange,fill=white,inner sep=2pt]
    at (axis cs:237,58) {$+19.7$ pp};

\end{groupplot}
\end{tikzpicture}
\caption{\textbf{Compact students perform well, with the largest stage benefit at 200 demonstrations.}
(a) Constrained selection improves the reasoning-trained student; the action-only configuration achieves higher success still.
(b) TPD leads C most clearly at 200 demonstrations; C catches up in mean unseen success at 404 and 808.
Error bars show sample standard deviations across three training seeds, and faint points show individual B/C runs.}

\label{fig:learning}
\end{figure}
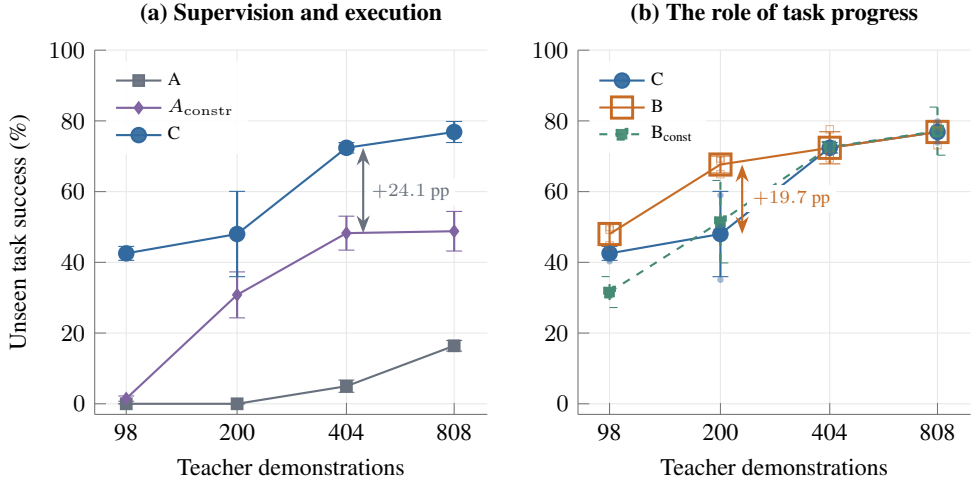

\paragraph{Constrained execution and the remaining performance gap.}
We first hold the reasoning-trained student's weights fixed and change how it selects actions.
At 404 demonstrations, A reaches 5.0\% success and frequently produces outputs that fail parsing.
Using the same checkpoints with \acon{} raises success to 48.3\%, a gain of 43.3 points.
At 98 demonstrations, about 99\% of A's steps lack a parseable action, and all episodes reach the 50-step limit without success. Constrained selection raises success to only 1.5\% at this budget. Changing the execution interface alone is therefore insufficient for this checkpoint configuration.

At the same 404-demonstration budget, C and TPD each exceed \acon{} by 24.1 percentage points.
At 808 demonstrations, the same separation holds: A reaches 16.4\%, \acon{} reaches 48.8\%, and both compact variants reach 76.9\%.
More demonstrations improve A from 5.0\% to 16.4\%, although its reported parse-failure rate remains 9.5--11.3\% across seeds at 808 demonstrations. It also records about 12,000 output tokens per episode, more than the other configurations.

\paragraph{Output volume during execution.}
At 404 demonstrations, TPD averages 22.0 steps and 299 output tokens per episode, compared with 33.6 steps and 9,995 tokens for \acon{} (Table~\ref{tab:cost}).
C averages 22.2 steps and 213 tokens.
The ratio of \acon{} to TPD output tokens is 33.4:1.
These counts describe trajectory text; they exclude the computation required to score B's $3N$ or C's $N$ candidates at each decision.

\begin{table}[t]
\centering\small
\caption{Execution at 404 demonstrations on unseen tasks.
Output totals are divided by 402 episodes; mean steps are averaged from the reported per-seed means.
Token counts exclude the additional work of scoring candidates.}

\label{tab:cost}
\begin{tabular}{@{\hspace{\tabcolsep}}lrrr@{\hspace{\tabcolsep}}}
\toprule
Arm & Success (\%) & Steps / episode & Output tokens / episode \\
\midrule
A & 5.0 & 47.9 & 14,635 \\
\acon & 48.3 & 33.6 & 9,995 \\
\rowcolor{ExpPaleOrange}
B / \tpd & \textbf{72.4} & 22.0 & 299 \\
\rowcolor{ExpPaleBlue}
C & \textbf{72.4} & 22.2 & 213 \\
\bottomrule
\end{tabular}
\end{table}

\paragraph{Reported results and training resources.}
Figure~\ref{fig:external} plots reported success against demonstration count for selected offline-supervised agents.
LM-BUTLER reaches 95\% with a 345M model and 3,553 expert demonstrations \cite{lmbutler}.
ETO's SFT stage reaches 67.2\% with a 7B model and 3,321 demonstrations containing expert actions and post-hoc reasoning annotations \cite{eto}.
Our compact students reach 72.4\% with a 1.7B model and 404 LLM-teacher demonstrations.
These points show the model and demonstration sizes associated with the reported results; training and evaluation protocols differ across papers.
Appendix~\ref{app:external-comparison} lists the resource details and includes methods whose actual demonstration counts remain unresolved.
Appendix~\ref{app:external-scaling} adds the HSL paper's SFT results at 800 and 1,600 demonstrations \cite{hsl}.

\subsection{Stage Benefits across Demonstration Budgets}
\label{sec:budget}
The action-only student already performs well with 404 demonstrations.
We next compare B and C across all four budgets to measure the added benefit of task stages (Figure~\ref{fig:learning}(b)).
The largest observed gain is at 200 demonstrations: B reaches 67.7\% mean unseen success and C reaches 48.0\%, a difference of 19.7 percentage points.
All three seeds favor B, by 29.9, 19.4, and 9.7 points for seeds 42, 1234, and 2026.

At 98 demonstrations, the gain is smaller at 5.5 points; at 404 and 808, the mean gap is zero.
Among the tested budgets, stages provide the largest additional benefit at 200 demonstrations; C matches B's mean unseen success with more action demonstrations.

The paired outcomes make the difference concrete (Table~\ref{tab:paired}).
At 200 demonstrations, B alone succeeds in 89 of the 402 seed--task evaluations, whereas C alone succeeds in ten.
At 808 demonstrations, each alone succeeds in 25 cases.
Equal mean success at 808 demonstrations therefore does not mean that the two policies succeed on exactly the same evaluations.

\begin{table}[t]
\centering\small
\caption{Paired B/C unseen outcomes over 134 tasks evaluated with three training seeds (402 seed--task records per budget).}
\label{tab:paired}
\begin{tabular}{@{}rrrrr@{}}
\toprule
Demonstrations & Both succeed & B only & C only & Both fail \\
\midrule
200 & 183 & 89 & 10 & 120 \\
808 & 284 & 25 & 25 & 68 \\
\bottomrule
\end{tabular}
\end{table}

On the seen diagnostic set, B leads C by 6.0, 13.8, 7.2, and 2.0 percentage points at 98, 200, 404, and 808 demonstrations, respectively.
The catch-up result refers specifically to unseen means.
We next test whether the stage benefit depends on labels that match the current task progress.

\subsection{The Value of Aligned Stage Labels}
\label{sec:alignment}
\paragraph{Constant-stage supervision.}
B adds a stage field to C's action target and changes candidate scoring to include that field.
We first replace every stage value with the constant word \texttt{phase}, giving \bconst{}.
At 200 demonstrations, this model reaches 51.5\% unseen success, compared with 67.7\% for B and 48.0\% for C.
The constant field does not reproduce B's advantage at this budget.
At 98 demonstrations, \bconst{} reaches 31.6\%, below B and C; at 404 and 808, the three means are close.

B and \bconst{} have identical target lengths for every 200-demonstration record: each stage value occupies one token, and the mean completion length is 13.69 tokens.
The constant control also removes label variation, changes the stage instruction, and reduces the candidate set from $3N$ to $N$.
The shuffled-stage control keeps all three labels and B's candidate structure, but changes the labels assigned to individual histories.

\paragraph{Stage alignment during training.}
\bshuffle{} permutes stage labels among decision records within each task type.
Histories, demonstrated actions, stage counts, and the $3N$ candidate structure remain unchanged.
All three training seeds use the same shuffled dataset, constructed with permutation seed 20261006.
Because labels can be permuted onto records with the same original value, 62.55\% of shuffled labels still match the originals.

Across the 116 seen tasks, B averages 55.2\% success and \bshuffle{} averages 37.6\% (Table~\ref{tab:shuffle}). The difference is 17.5 percentage points when computed from unrounded success counts.
B leads in all three seeds; C and \bconst{} reach 41.4\% and 43.7\%.
Keeping the label vocabulary and candidate structure is therefore insufficient to retain B's seen-set performance when the labels are shuffled.
The result supports using labels that match the current task progress.
The comparison also includes the greater difficulty of predicting shuffled labels from the history.

\begin{table}[t]
\centering
\small
\caption{Stage-alignment controls at 200 demonstrations: success (\%) on the 116 seen diagnostic tasks.
Results are shown for three training seeds and their mean; bold marks the highest value in each column.}

\label{tab:shuffle}
\begin{tabular}{@{\hspace{\tabcolsep}}lrrrr@{\hspace{\tabcolsep}}}
\toprule
Arm & Seed 42 & Seed 1234 & Seed 2026 & Mean \\
\midrule
\rowcolor{ExpPaleOrange}
B / \tpd & \textbf{55.2} & \textbf{53.4} & \textbf{56.9} & \textbf{55.2} \\
\rowcolor{ExpPaleBlue}
C & 35.3 & 37.9 & 50.9 & 41.4 \\
\bconst & 31.9 & 49.1 & 50.0 & 43.7 \\
\bshuffle & 31.0 & 42.2 & 39.7 & 37.6 \\
\bottomrule
\end{tabular}
\end{table}

Appendix~\ref{app:shuffle-diagnostics} reports the additional open-loop diagnostics and replay-annotation coverage.

\subsection{Task Progress and the Next Decision}
\label{sec:mechanism}
At 200 demonstrations, B outperforms \bconst{} on unseen tasks and \bshuffle{} on seen tasks.
We now examine what the agent does differently.
We first locate behavioral differences in complete trajectories, then hold the history fixed to compare the next action.

\paragraph{Starting processing after object acquisition.}
At 200 demonstrations, B's unseen gain over C is largest for heating (+36.2 percentage points), followed by simple picking (+22.2) and cleaning (+20.4).
The gains for picking two objects, looking under light, and cooling are 13.7, 13.0, and 7.9 points.
We use environment feedback to identify when an agent acquires the target, reaches a facility, and performs the required processing.

The heating trajectories show where the policies begin to differ after acquisition.
After acquiring the correct object, B reaches the microwave in 56/58 cases, compared with 32/56 for C.
It successfully heats the object in 55/58 cases, compared with 30/56 for C.
Neither arm had reached the microwave before acquisition in these episodes.
In these episodes, a substantial difference appears between acquiring the object and initiating the required processing.

We then restrict the comparison to seed--task pairs in which both agents acquired the correct object (Table~\ref{tab:processing}).
B completes heating in 49/52 records and C in 28/52.
Cleaning shows a smaller gap, and cooling shows little difference.
All heating and cleaning attempts in these cohorts succeed when issued, so the main difference at this point is whether the agent initiates processing.
The two agents may still have different histories and remaining steps when they acquire the object. We address this difference by comparing actions on shared histories below.

\begin{table}[t]
\centering\small
\caption{Successful processing after correct acquisition, restricted to pairs in which both agents acquire the target (200 demonstrations, unseen). Task and training seed are paired; the acquisition-time environment states are not identical.}
\label{tab:processing}
\begin{tabular}{@{}lrrr@{}}
\toprule
Task & Paired records & B processes successfully & C processes successfully \\
\midrule
Heat & 52 & \textbf{49/52 (94.2\%)} & 28/52 (53.8\%) \\
Clean & 60 & \textbf{59/60 (98.3\%)} & 51/60 (85.0\%) \\
Cool & 49 & 44/49 (89.8\%) & \textbf{45/49 (91.8\%)} \\
\bottomrule
\end{tabular}
\end{table}

The policies also differ before processing begins.
C also more often picks up the wrong type of object across several task families.
In failed simple-picking episodes, the target never appears in the observation history in 11/24 C cases and 3/8 B cases.
Loops of at least five identical consecutive actions occur at similar overall rates (22.6\% for B and 21.9\% for C), but in 90/130 B failures and 88/209 C failures.
The gap therefore includes both failure to find or acquire the right object and failure to move on to processing it.

\paragraph{Comparing actions on shared histories.}
We evaluate B and C on 72 frozen seen prefixes from 50 tasks.
Each prefix ends just before an action, giving both models the same interaction history and admissible actions; only their output-format instructions differ.
We test the 200-, 404-, and 808-demonstration checkpoints for all three seeds.
The pool contains 24 pilot prefixes and 48 additional prefixes sampled without consulting the new scores.

We use two tests to examine how B's stage output affects action selection.
First, we keep the stage in the candidate text but sum only the action-value token scores, excluding stage and JSON-structure scores.
We call this \emph{action-value-only} scoring.
Second, we fix each possible stage in turn and compare the highest-scoring actions under those stage conditions.
The first test changes which tokens contribute to the score; the second changes the stage on which the action is conditioned.

Removing the non-action score terms changes B's selected action in 86/648 evaluations (13.3\%), across 72 prefixes, three budgets, and three seeds.
Fixing different stage values changes the top action in 130/648 (20.1\%).
A blinded review of 36 actions changed by score masking judges 11 changes helpful, six harmful, and 19 undetermined.
Both the stage condition and the score terms can change a choice; the blinded review shows why a changed choice need not be a better one.
On the same prefix set, B/C action agreement rises from about 83\% at 200 demonstrations to about 90\% at 404 and 808 across the three seeds.

\paragraph{Local progress under fixed stage conditions.}
We next test whether matching the stage to the task's progress improves the selected action.
For 38 prefixes from 30 tasks, the reference stage is reliable and the rules identify at least one admissible action that makes progress.
Using only the current prefix, the rules classify candidate actions as \textsc{Progress}, \textsc{Counterproductive}, or \textsc{Undetermined}.
For each B checkpoint, we select an action with the matching stage fixed, then repeat with each of the two mismatched stages.
All conditions use action-value-only scoring.

At 200 demonstrations, fixing the matching stage yields a progress action in 83.3\% of evaluations, compared with 65.4\% averaged over the two mismatched stages.
Computed from unrounded counts, the gap is 18.0 percentage points, with positive differences for all three seeds (14.5, 18.4, and 21.1 points). Giving each task equal weight yields a 16.3-point gap.
The clearest differences occur in heating and some visible-target acquisition decisions, while cleaning and cooling subsets are already near saturation.
These rules cover identifiable progress actions, but do not evaluate how well the agent searches for targets it has not yet observed.

\begin{figure}[t]
\centering
\begin{subfigure}[t]{0.47\textwidth}
\centering
\vspace{0pt}
\fbox{\begin{minipage}{0.91\linewidth}
\small
\textbf{Shared history: heating a cup}\\[3pt]
The task requires a heated cup.\\
The agent has just acquired the cup.\\[7pt]
\textbf{Next decision:}\par
\textcolor{TPDOrange}{TPD, 200 demos:}\\
Microwave selected: \textbf{3/3 seeds}.\\[5pt]
\textcolor{ActionBlue}{Action-only, 200 demos:}\\
Microwave selected: \textbf{1/3 seeds};\\
further search: 2/3.\\[5pt]
\textcolor{ActionBlue}{Action-only, 404/808 demos:}\\
Microwave selected: \textbf{6/6 checkpoints}.\\[7pt]
\footnotesize One illustrative decision across training seeds and budgets.
\end{minipage}}
\caption{Choosing the microwave after acquiring the cup.}
\end{subfigure}\hfill
\begin{subfigure}[t]{0.50\textwidth}
\centering\vspace{0pt}
\begin{tikzpicture}
\begin{axis}[width=\linewidth,height=6.1cm,xmode=log,log basis x=2,xmin=175,xmax=925,
 ymin=0,ymax=100,xtick={200,404,808},xticklabels={200,404,808},
 xlabel={Demonstrations},ylabel={Local progress rate (\%)},
 tick label style={font=\small},label style={font=\small},
 grid=major,grid style={black!8},legend pos=south east,
 legend style={font=\scriptsize,draw=none},mark size=2.5pt]
\addplot[TPDOrange,thick,mark=*] coordinates {(200,83.3)(404,82.5)(808,78.1)};
\addlegendentry{Matching stage}
\addplot[ReasonGray,thick,dashed,mark=square*] coordinates {(200,65.4)(404,75.4)(808,75.4)};
\addlegendentry{Mismatched stages (average)}
\end{axis}
\end{tikzpicture}
\caption{Local progress under fixed stage conditions.}
\end{subfigure}
\caption{\textbf{Moving from object acquisition to processing.}
(a) On the same history after acquiring a cup, B chooses the microwave at all nine checkpoints.
C chooses it at one of three 200-demonstration checkpoints and all six larger-budget checkpoints.
(b) Across 38 rule-evaluable seen prefixes from 30 tasks, fixing the matching stage more often yields a progress action than fixing a mismatched stage.
Curves show local progress rates averaged over three seeds at each budget.
Table~\ref{tab:overlap} relates these tests to the original B/C choices.}

\label{fig:mechanism}
\end{figure}
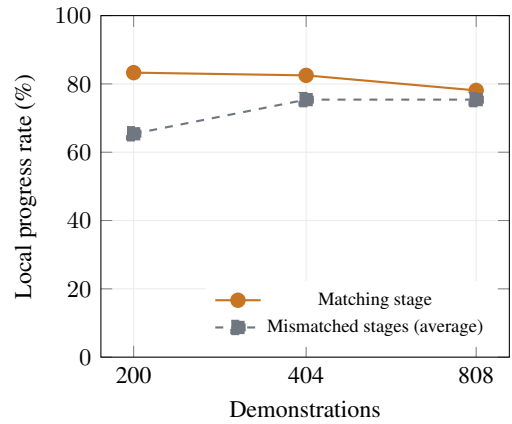

Figure~\ref{fig:mechanism} illustrates the decision change with a shared history from the cup task.
On the same history immediately after acquisition, all three B checkpoints at 200 demonstrations choose the microwave, while two of the three C checkpoints continue searching.
All six C checkpoints at 404 and 808 demonstrations choose the microwave.
Across the 38 evaluable prefixes, the matching-stage advantage narrows as demonstrations increase: progress under mismatched stages rises, while progress under the matching stage falls.
The example shows C learning to make a choice that B already makes at 200 demonstrations; the aggregate test shows a smaller effect of fixing the stage at larger budgets.

\begin{table}[t]
\centering\small
\caption{Local progress and stage sensitivity at 200 demonstrations: 38 seen prefixes evaluated with three training seeds.
Overlap fractions use the 15 B-only progress cases as their denominator.
Coverage and scoring checks show how many tasks support the overlap and whether it persists under both scoring rules.}

\label{tab:overlap}
\begin{tabularx}{\textwidth}{@{}Yr@{}}
\toprule
Criterion or coverage & Result \\
\midrule
Both original policies choose progress & 81/114 \\
Only original B chooses progress & 15/114 \\
Only original C chooses progress & 1/114 \\
Neither original policy chooses progress & 17/114 \\
\midrule
B-only cases with useful matching stage and sensitivity to a mismatch & 11/15 \\
Also require original B to select reference stage and matching action & 11/15 \\
Also require full-joint and global action-value-only actions to agree & 2/15 \\
Unique coverage of the 11 strictly aligned cases & 5 prefixes; 3 tasks \\
Heating/transform cases within those 11 & 9/11 \\
\bottomrule
\end{tabularx}
\end{table}

\paragraph{Overlap with B's local advantage.}
The fixed-stage tests show which action B selects when we supply a stage.
We now compare B's normal joint selection with C's action-only selection on the same 38 prefixes.
At 200 demonstrations, both choose a progress action in 81 of the 114 prefix--seed evaluations, only B in 15, only C in one, and neither in 17.
B's local progress rate is 84.2\%, compared with 71.9\% for C; the three seeds contribute five, six, and four B-only cases.

In 11 of those 15 B-only cases, the matching-stage action makes progress and at least one mismatched-stage action does not.
In all eleven cases, B's normal joint selection already chooses the matching stage and the same progress action.
Part of B's local advantage therefore overlaps with decisions that are sensitive to the stage condition.
The eleven cases come from five prefixes and three tasks; nine cases involve heating at the transform stage.
Only two of the eleven cases preserve B's selected action when switching from full-joint to global action-value-only scoring (Table~\ref{tab:overlap}). The overlap is therefore sensitive to the scoring rule.

Useful stage sensitivity occurs in 73\% of B-only cases, 21\% of cases where both policies select progress, and 6\% of cases where neither does.
In 16 of the 17 cases where neither policy selects progress, B still selects a non-progress action when the matching stage is fixed.
Fixing the stage alone therefore does not resolve these local failures.
Across 200, 404, and 808 demonstrations, B-only cases decrease from 15 to seven to two, and stage-sensitive cases from 29 to 16 to nine.

The results suggest that task-stage supervision helps with some transitions between subgoals and that the stage condition can guide the next action at those points.
With more demonstrations, C agrees more often with B on the sampled histories, while their mean unseen success rates converge.
These local tests identify decisions consistent with this explanation; determining their contribution to episode success would require closed-loop interventions.

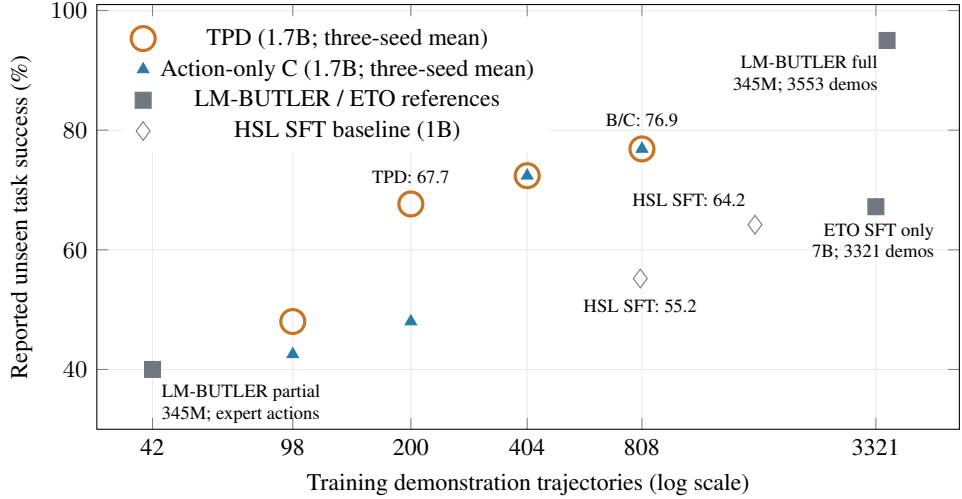
\begin{figure}[t]
\centering
\begin{tikzpicture}
\begin{axis}[width=.93\textwidth,height=7.2cm,
 xmode=log,xmin=30,xmax=5500,ymin=30,ymax=101,
 xtick={42,98,200,404,808,3321},xticklabels={42,98,200,404,808,3321},
 xlabel={Training demonstration trajectories (log scale)},
 ylabel={Reported unseen task success (\%)},
 grid=major,grid style={black!8},tick label style={font=\small},
 label style={font=\small},legend pos=north west,
 legend style={font=\small,draw=none,fill=white},clip=false]
\addplot[only marks,TPDOrange,mark=o,mark size=4.5pt,very thick] coordinates {(98,48.010)(200,67.662)(404,72.388)(808,76.866)};
\addlegendentry{TPD (1.7B; three-seed mean)}
\addplot[only marks,ActionBlue,mark=triangle*,mark size=2.5pt] coordinates {(98,42.537)(200,48.010)(404,72.388)(808,76.866)};
\addlegendentry{Action-only C (1.7B; three-seed mean)}
\addplot[only marks,ReasonGray,mark=square*,mark size=3pt] coordinates {(42,40)(3553,95)(3321,67.2)};
\addlegendentry{LM-BUTLER / ETO references}
\addplot[only marks,ReasonGray,mark=diamond,mark size=3.5pt] coordinates {(800,55.2)(1600,64.2)};
\addlegendentry{HSL SFT baseline (1B)}
\node[anchor=north west,font=\scriptsize,align=left] at (axis cs:42,39) {LM-BUTLER partial\\345M; expert actions};
\node[anchor=north east,font=\scriptsize,align=right] at (axis cs:3553,94) {LM-BUTLER full\\345M; 3553 demos};
\node[anchor=north east,font=\scriptsize,align=right] at (axis cs:5000,66) {ETO SFT only\\7B; 3321 demos};
\node[anchor=south,font=\scriptsize] at (axis cs:200,69.5) {TPD: 67.7};
\node[anchor=south,font=\scriptsize] at (axis cs:808,79) {B/C: 76.9};
\node[anchor=north,font=\scriptsize] at (axis cs:800,53.2) {HSL SFT: 55.2};
\node[anchor=south east,font=\scriptsize] at (axis cs:1600,65.8) {HSL SFT: 64.2};
\end{axis}
\end{tikzpicture}
\caption{\textbf{Reported success and demonstration counts
for offline-supervised agents.}
External points show LM-BUTLER, ETO's SFT stage, and the
800- and 1,600-demonstration SFT baselines reported in HSL
\cite{lmbutler,eto,hsl}.
Our B/C results are three-seed means; their markers overlap
at 404 and 808 demonstrations.
Each work uses its own student model, demonstration source,
and evaluation protocol.
Appendix~\ref{app:external-scaling} lists the plotted values
and their sources.}

\label{fig:external}
\end{figure}

\section{Limitations}
\label{sec:limitations}
The experiments cover one environment and one student model size.
ALFWorld supplies admissible actions, and the student receives the full interaction history.
The annotator supplies task knowledge through hand-defined progress rules.
Using the method in another environment would require suitable progress rules and an interface for constructing admissible actions.

Several factors change together in the comparisons.
Reasoning and compact configurations differ in target length, the fraction of action tokens, and the reasoning context used by \acon{} during action selection.
The constant-stage control changes candidate count, and shuffling changes label predictability.
Keeping training at three epochs means that larger datasets increase both demonstration coverage and training computation.
Preserving task-category proportions does not control the examples or difficulty within each category.
The current results do not separate these contributions.

The three seeds measure training variation on one fixed dataset at each budget.
The seen set is filtered by expert success and includes the checkpoint-selection probe, whereas unseen evaluation uses the full split.
The shuffle and shared-history analyses are exploratory studies on this seen distribution.
The overlap between stage sensitivity and B-only progress covers three tasks and changes with the scoring rule.
Measuring its contribution to episode success would require a closed-loop intervention.

Output-token counts measure trajectory text and exclude candidate-scoring computation.
End-to-end latency and total training cost remain to be measured.

\section{Conclusion}
\label{sec:conclusion}
We train small agents to execute multi-step tasks from fixed teacher demonstrations using compact decision targets and admissible-action scoring.
On ALFWorld, both the action-only student and TPD reach 72.4\% mean unseen success with 404 demonstrations and 76.9\% with 808 using a 1.7B model.
TPD adds its largest observed benefit at 200 demonstrations, raising success from 48.0\% to 67.7\%; the action-only student matches its mean unseen success at the two larger budgets.
Shared-history tests link some of this local advantage to decisions that advance the task, such as moving to the microwave after acquiring an object that needs heating.
These results establish compact action supervision as a useful baseline and show when explicit task progress adds value in this setting.


\appendix

\section{Implementation and Reproducibility Details}
\label{app:implementation-details}

\paragraph{Stage annotation.}
The annotator is defined in version 2 of \path{configs/fsm.yaml} and implemented in \path{src/annotate_states.py}.
It uses the task type and three tracked facts: whether the target is held, whether processing is complete, and whether placement has occurred.
After labeling the current decision, it updates these facts from recognized success messages in the environment feedback.
Heating, cleaning, and cooling require a transformation.
The rules use a fixed object vocabulary and target-inference logic to identify the target. Unsupported action verbs trigger an assertion.

\paragraph{Training recipe.}
We fine-tune all parameters using bfloat16, scaled dot-product attention, and gradient checkpointing, without LoRA.
We use fused PyTorch AdamW with a learning rate of $10^{-5}$, cosine decay, 10 warmup steps, and zero weight decay.
The microbatch size is eight, with four gradient-accumulation steps and effective batch size 32.
Sequences are limited to 8,192 tokens.
We train for three epochs and evaluate the seen probe after each epoch.
The epoch with the highest action agreement is selected; ties favor the earlier epoch.
The probe has six tasks from each of five task families and five picking-two tasks, totaling 697 decisions from 35 tasks.

\paragraph{Loss and serialization.}
The loss covers the completion, with system and user tokens masked out.
For TPD, this includes the stage and action values, JSON syntax, the empty thinking block, and the chat-template end-of-message token.
There are no separate loss weights for stage and action tokens.
Training logs record aggregate completion loss, without separate stage and action loss curves.
Dataset construction checks assistant targets against the prompt templates.
Candidate scoring uses the same compact serialization. Prefix-reconstruction checks verify that the scoring inputs match the intended histories.

\paragraph{Candidate order.}
The outer loop follows the environment's admissible-action order, and the inner loop follows the fixed stage order.
The first candidate wins a tie.
Scores are sums of completion-token log probabilities, with no length normalization.

\paragraph{Evaluation subset.}
The rule-based expert completes 116 of the 140 official seen tasks within 50 steps; all 116 form the seen diagnostic set.
It contains 35 simple-picking, 26 cleaning, 24 cooling, 14 heating, 12 looking-under-light, and five picking-two tasks.
The 35 checkpoint-probe tasks are a subset of this set.
Unseen evaluation uses all 134 official tasks, with no expert-success filtering.

\paragraph{Software and model versions.}
The recorded environment uses ALFWorld 0.4.2, TextWorld 1.7.0, PyTorch 2.11.0, Transformers 5.17, and TRL 1.13.0.
Student inference uses a local Hugging Face Transformers service.
The common initialization is the post-trained \texttt{Qwen/Qwen3-1.7B} snapshot at revision:
\begin{quote}\small\texttt{70d244cc86ccca08cf5af4e1e306ecf908b1ad5e}\end{quote}
Teacher collection uses \texttt{deepseek-v4-pro}, the \texttt{teacher\_v2} prompt, temperature 0.7, and full interaction history.

\section{Shuffled-Stage Diagnostic Protocol}
\label{app:shuffle-diagnostics}

\paragraph{Free-generation and constrained-execution legality.}
At seed 42, the shuffled model produces legal outputs on 79.5\% of 1,870 open-loop records when evaluated by free generation.
An output is counted as legal if it can be parsed, its stage is allowed, and its action belongs to the admissible set.
The shuffled rates at seeds 1234 and 2026 are 82.5\% and 84.0\%; B's rates at seeds 42, 1234, and 2026 are 84.1\%, 83.2\%, and 84.5\%.

Closed-loop execution selects from candidates built with allowed stages and admissible actions.
Those selected actions are legal by construction.
The free-generation diagnostic measures how often the model produces such an output without candidate selection.

\paragraph{Replay annotation coverage.}
At seed 42, the annotator can label 76/116 B trajectories and 51/116 shuffled-model trajectories.
Both can be labeled on 47 tasks, only B on 29, only the shuffled model on four, and neither on 36.

In these runs, the annotator fails when the trajectory never successfully operates on the target object.
The annotatable sets therefore depend on the policies' behavior.
The common set for a paired replay comparison contains 47 tasks, whereas the success rates in Table~\ref{tab:shuffle} use all 116.

\paragraph{Interpretation of open-loop diagnostics.}
At seed 42, the shuffled model and B have similar open-loop action agreement (49.1\% and 48.4\%), despite their different closed-loop success rates.
The open-loop test supplies the history and compares one action with a reference.
In closed-loop execution, each policy must also handle the histories created by its earlier choices.

The shuffled model's 71.7\% agreement with reference stages is measured during evaluation.
The 62.55\% agreement between shuffled and original labels describes the training data.
Because these percentages use different data and denominators, their difference does not measure recovery of stage knowledge.

\clearpage
\section{Comparison with Published Agent-Training Results}
\label{app:external-comparison}

Table~\ref{tab:external-resources} lists the training resources and evaluation protocols of the methods discussed in the main text.
It includes methods whose actual demonstration counts remain unresolved and are therefore absent from Figure~\ref{fig:external}.

\par\vspace{8pt}
\begingroup
\centering
\captionof{table}{Training resources and reported ALFWorld results under each work's evaluation protocol.
Our values are three-seed means; standard deviations appear in Table~\ref{tab:main}.
Trajectory counts and decision-record counts are listed separately.}

\label{tab:external-resources}
\vspace{4pt}
\fontsize{8.5}{10.5}\selectfont
\setlength{\tabcolsep}{4pt}
\renewcommand{\arraystretch}{1.22}
\begin{tabularx}{\linewidth}{@{}p{112pt}p{78pt}rrXcX@{}}
\toprule
 & & \multicolumn{3}{c}{Training resources} & Evaluation & Success \\
\cmidrule(lr){3-5}
Method & Student & Demos & Demo steps & Interaction & protocol & (\%) \\
\midrule
\multicolumn{7}{@{}l}{\rule{0pt}{2.4ex}\textit{Offline SFT / imitation: external reported results}}\\[2pt]
LM-BUTLER \cite{lmbutler} & GPT2-medium (345M) & 42 & \nr & No & Original paper & 40.0 \\
LM-BUTLER \cite{lmbutler} & GPT2-medium (345M) & 3,553 & \nr & No & Original paper & 95.0 \\
ETO: SFT stage \cite{eto} & Llama-2-7B-Chat & 3,321 & \nr & No & Original paper & 67.2 \\
HSL: SFT baseline \cite{hsl} & Llama3.2-1B-Instruct & $>$3,200 & \nr & No & Original paper & 78.36 \\
SFT \cite{opcsft} & Qwen2.5-1.5B & \uv & \uv & No & EMBod-Bench & 70.90 \\
OPC-SFT \cite{opcsft} & Qwen2.5-1.5B & \uv & \uv & No & EMBod-Bench & 72.39 \\
SFT \cite{opcsft} & Qwen2.5-1.5B & \uv & \uv & No & AgentBoard & 73.88 \\
OPC-SFT \cite{opcsft} & Qwen2.5-1.5B & \uv & \uv & No & AgentBoard & 76.12 \\
\addlinespace[3pt]
\midrule
\multicolumn{7}{@{}l}{\rule{0pt}{2.4ex}\textit{Additional environment-interaction training: external references}}\\[2pt]
ETO \cite{eto} & Llama-2-7B-Chat & 3,321 & \nr & Yes & Original paper & 72.4 \\
SFT + HSL \cite{hsl} & Llama3.2-1B-Instruct & $>$3,200 & \nr & Yes & Original paper & 97.76 \\
\addlinespace[3pt]
\midrule
\multicolumn{7}{@{}l}{\rule{0pt}{2.4ex}\textit{This work: offline SFT with constrained candidate selection}}\\[2pt]
Action-only (C) & Qwen3-1.7B & 200 & 3,712 & No & Our harness & 48.0 \\
TPD (B) & Qwen3-1.7B & 200 & 3,712 & No & Our harness & 67.7 \\
Action-only (C) & Qwen3-1.7B & 404 & 7,007 & No & Our harness & 72.4 \\
TPD (B) & Qwen3-1.7B & 404 & 7,007 & No & Our harness & 72.4 \\
Action-only (C) & Qwen3-1.7B & 808 & 13,618 & No & Our harness & 76.9 \\
TPD (B) & Qwen3-1.7B & 808 & 13,618 & No & Our harness & 76.9 \\
\bottomrule
\end{tabularx}
\endgroup
\vspace{7pt}
\begingroup
\fontsize{8}{10}\selectfont
\setlength{\parindent}{0pt}
\setlength{\parskip}{3.5pt}
\noindent{\color{black!25}\rule{\linewidth}{0.4pt}}\par\vspace{2pt}
\textit{Notes.}
Demos counts demonstration trajectories; Demo steps counts individual decision records.
Student interaction records whether training uses environment rollouts beyond the fixed demonstrations; teacher collection and evaluation are excluded.
The demonstration counts for interaction-based methods describe
their initial supervised data and do not measure total training
experience.
The SFT + HSL method \cite{hsl} also uses a 70B relabeler; its SFT-only baseline does not include this additional stage.

External methods differ in teacher source, model, checkpoint selection, prompts, and execution.
Our results cover all 134 \texttt{valid\_unseen} tasks with a 50-step limit.
External values retain their published precision.
Table~\ref{tab:external-scaling} provides the full four-budget
B/C comparison.
The remaining internal arms are reported in the main-text tables.

\textit{Abbreviations.}
\nr: not reported in the cited source.
\uv: actual training-data usage has not been verified. For OPC-SFT \cite{opcsft}, the available data-volume accounts do not establish the amount used for each reported run.

\textit{Source locations.}
LM-BUTLER: Table~1 \cite{lmbutler};
ETO: Tables~1 and~7 \cite{eto};
OPC-SFT: Tables~1 and~9, before RL \cite{opcsft};
HSL: Table~1 and Figure~3 \cite{hsl}.
\endgroup

\clearpage
\subsection{Reported Performance across Demonstration Budgets}
\label{app:external-scaling}

A method's full-data result does not show how well it learns
from fewer demonstrations.
Table~\ref{tab:external-scaling} compares the smaller-budget SFT results reported in HSL \cite{hsl} with LM-BUTLER \cite{lmbutler}, ETO's SFT stage \cite{eto}, and our four demonstration budgets.
Table~\ref{tab:external-resources} provides the corresponding
training-resource and evaluation details.

\par\vspace{8pt}
\begingroup
\centering
\captionof{table}{Reported ALFWorld unseen success across
demonstration budgets.
External entries use each paper's evaluation protocol.
Our entries are means over three training seeds; their
standard deviations appear in Table~\ref{tab:main}.}
\label{tab:external-scaling}
\vspace{4pt}

\small
\setlength{\tabcolsep}{5pt}
\renewcommand{\arraystretch}{1.16}

\begin{tabularx}{\linewidth}{@{}XXrrl@{}}
\toprule
Method & Student & Demos & Success (\%) & Source \\
\midrule

\multicolumn{5}{@{}l}{
  \rule{0pt}{2.4ex}
  \textit{Published offline-supervised results}
}\\[2pt]

LM-BUTLER, partial \cite{lmbutler}
& GPT2-medium (345M)
& 42
& 40.0
& Table 1 \\

LM-BUTLER, full \cite{lmbutler}
& GPT2-medium (345M)
& 3,553
& 95.0
& Table 1 \\

\addlinespace[3pt]

HSL: SFT baseline \cite{hsl}
& Llama3.2-1B-Instruct
& 800
& 55.2
& Figure 3 \\

HSL: SFT baseline \cite{hsl}
& Llama3.2-1B-Instruct
& 1,600
& 64.2
& Figure 3 \\

HSL: SFT baseline \cite{hsl}
& Llama3.2-1B-Instruct
& $>3{,}200$
& 78.36
& Table 1 \\

\addlinespace[3pt]

ETO: SFT stage \cite{eto}
& Llama-2-7B-Chat
& 3,321
& 67.2
& Tables 1, 7 \\

\midrule

\multicolumn{5}{@{}l}{
  \rule{0pt}{2.4ex}
  \textit{This work: offline SFT with constrained action selection}
}\\[2pt]

Action-only (C)
& Qwen3-1.7B
& 98
& 42.5
& Table~\ref{tab:main} \\

TPD (B)
& Qwen3-1.7B
& 98
& 48.0
& Table~\ref{tab:main} \\

\addlinespace[3pt]

Action-only (C)
& Qwen3-1.7B
& 200
& 48.0
& Table~\ref{tab:main} \\

TPD (B)
& Qwen3-1.7B
& 200
& 67.7
& Table~\ref{tab:main} \\

\addlinespace[3pt]

Action-only (C)
& Qwen3-1.7B
& 404
& 72.4
& Table~\ref{tab:main} \\

TPD (B)
& Qwen3-1.7B
& 404
& 72.4
& Table~\ref{tab:main} \\

\addlinespace[3pt]

Action-only (C)
& Qwen3-1.7B
& 808
& 76.9
& Table~\ref{tab:main} \\

TPD (B)
& Qwen3-1.7B
& 808
& 76.9
& Table~\ref{tab:main} \\

\bottomrule
\end{tabularx}
\par
\endgroup

\vspace{5pt}
\begingroup
\footnotesize
\setlength{\parindent}{0pt}

\textit{Notes.}
Demos counts training demonstration trajectories.
All entries report offline-supervised performance before any additional student exploration or online optimization.
The HSL rows refer to its SFT baseline, not the full HSL method \cite{hsl}.
HSL's smaller-budget values come from Figure~3 and its full-budget score from Table~1; the full demonstration count is reported only as $>3{,}200$.
Methods with unresolved actual demonstration counts are omitted.
\par
\endgroup

\paragraph{Reference results at smaller budgets.}
With 808 demonstrations, both of our compact students reach 76.9\% mean unseen success; the HSL paper reports 55.2\% for its SFT baseline with 800 demonstrations \cite{hsl}.
TPD also reaches 67.7\% with 200 demonstrations, compared with
64.2\% for that baseline at 1,600 demonstrations.
The HSL SFT baseline \cite{hsl} reaches 78.36\% at its full budget of
more than 3,200 demonstrations.
These results provide reference points for learning from a few hundred demonstrations.
Because the studies differ in student models, demonstration sources, and execution protocols, these are comparisons of reported systems rather than isolated training methods.


\begin{thebibliography}{99}

\bibitem[Yang et~al.(2025)]{qwen3}
An Yang, Anfeng Li, Baosong Yang, Beichen Zhang, Binyuan Hui,
Bo Zheng, Bowen Yu, Chang Gao, Chengen Huang, Chenxu Lv, and others.
\newblock \emph{Qwen3 Technical Report}.
\newblock arXiv:2505.09388, 2025.
\newblock \url{https://arxiv.org/abs/2505.09388}

\bibitem[Shridhar et~al.(2021)]{alfworld}
Mohit Shridhar, Xingdi Yuan, Marc-Alexandre C\^ot\'e,
Yonatan Bisk, Adam Trischler, and Matthew Hausknecht.
\newblock \emph{{ALFWorld}: Aligning Text and Embodied
Environments for Interactive Learning}.
\newblock In \emph{International Conference on Learning
Representations (ICLR)}, 2021.
\newblock \url{https://github.com/alfworld/alfworld}


\bibitem[Micheli and Fleuret(2021)]{lmbutler}
Vincent Micheli and Fran\c{c}ois Fleuret.
\newblock \emph{Language Models are Few-Shot Butlers}.
\newblock In \emph{Proceedings of the 2021 Conference on
Empirical Methods in Natural Language Processing (EMNLP)},
pages 9312--9318. Association for Computational Linguistics,
2021.
\newblock \url{https://aclanthology.org/2021.emnlp-main.734/}

\bibitem[Chen et~al.(2023)]{fireact}
Baian Chen, Chang Shu, Ehsan Shareghi, Nigel Collier,
Karthik Narasimhan, and Shunyu Yao.
\newblock \emph{FireAct: Toward Language Agent Fine-tuning}.
\newblock arXiv:2310.05915, 2023.
\newblock \url{https://arxiv.org/abs/2310.05915}

\bibitem[Zeng et~al.(2024)]{agenttuning}
Aohan Zeng, Mingdao Liu, Rui Lu, Bowen Wang, Xiao Liu,
Yuxiao Dong, and Jie Tang.
\newblock \emph{AgentTuning: Enabling Generalized Agent
Abilities for LLMs}.
\newblock In \emph{Findings of the Association for
Computational Linguistics: ACL 2024}, pages 3053--3077.
Association for Computational Linguistics, 2024.
\newblock \url{https://aclanthology.org/2024.findings-acl.181/}

\bibitem[Liu et~al.(2025)]{sad}
Jun Liu, Zhenglun Kong, Peiyan Dong, Changdi Yang,
Tianqi Li, Hao Tang, Geng Yuan, Wei Niu, Wenbin Zhang,
Pu Zhao, Xue Lin, Dong Huang, and Yanzhi Wang.
\newblock \emph{Structured Agent Distillation for Large
Language Model Agents}.
\newblock arXiv:2505.13820, 2025.
\newblock \url{https://arxiv.org/abs/2505.13820}

\bibitem[Liu et~al.(2026)]{opcsft}
Tian-Shuo Liu, Chengxing Jia, Haoyu Liu, Pengyuan Wang,
Shiyuan Zhang, Jie Fu, and Yang Yu.
\newblock \emph{Clipping Low-Probability Tokens in {SFT}
Yields a Generalizable Initialization for {RL}}.
\newblock In \emph{Proceedings of the 43rd International
Conference on Machine Learning (ICML)}, volume 306 of
\emph{Proceedings of Machine Learning Research},
pages 76871--76896. PMLR, 2026.
\newblock \url{https://proceedings.mlr.press/v306/liu26bm.html}

\bibitem[Song et~al.(2024)]{eto}
Yifan Song, Da Yin, Xiang Yue, Jie Huang, Sujian Li, and
Bill Yuchen Lin.
\newblock \emph{Trial and Error: Exploration-Based Trajectory
Optimization of {LLM} Agents}.
\newblock In \emph{Proceedings of the 62nd Annual Meeting of
the Association for Computational Linguistics (Volume 1:
Long Papers)}, pages 7584--7600. Association for
Computational Linguistics, 2024.
\newblock \url{https://aclanthology.org/2024.acl-long.409/}

\bibitem[Yang et~al.(2024)]{a3t}
Zonghan Yang, Peng Li, Ming Yan, Ji Zhang, Fei Huang, and
Yang Liu.
\newblock \emph{ReAct Meets ActRe: Autonomous Annotation of
Agent Trajectories for Contrastive Self-Training}.
\newblock In \emph{First Conference on Language Modeling
(COLM)}, 2024.
\newblock \url{https://arxiv.org/abs/2403.14589}

\bibitem[Gao et~al.(2025)]{hpl}
Heyang Gao et al. 
\newblock \emph{Solving the Granularity Mismatch:
Hierarchical Preference Learning for Long-Horizon {LLM}
Agents}.
\newblock arXiv:2510.03253, 2025.
\newblock \url{https://arxiv.org/abs/2510.03253}

\bibitem[Zhou et~al.(2026)]{sageopd}
Yuhang Zhou, Lizhu Zhang, Yifan Wu, Mingyi Wang, Bo Peng,
Jiayi Liu, Xiangjun Fan, and Zhuokai Zhao.
\newblock \emph{{SAGE-OPD}: Selective Agent-Guided
Intervention for Multi-Turn On-Policy Distillation}.
\newblock arXiv:2606.19659, 2026.
\newblock \url{https://arxiv.org/abs/2606.19659}

\bibitem[Li et~al.(2026)]{hsl}
Zichao Li, Gang Wu, Zichao Wang, Ruiyi Zhang, Wanrong Zhu,
Ryan A. Rossi, Vlad I. Morariu, and Jihyung Kil.
\newblock \emph{Spinning Straw into Gold: Relabeling {LLM}
Agent Trajectories in Hindsight for Successful
Demonstrations}.
\newblock In \emph{International Conference on Learning
Representations (ICLR)}, 2026.
\newblock \url{https://arxiv.org/abs/2607.04235}

\bibitem[Feng et~al.(2025)]{gigpo}
Lang Feng, Zhenghai Xue, Tingcong Liu, and Bo An.
\newblock \emph{Group-in-Group Policy Optimization for {LLM}
Agent Training}.
\newblock In \emph{Advances in Neural Information Processing
Systems (NeurIPS)}, 2025.
\newblock \url{https://arxiv.org/abs/2505.10978}

\bibitem[Peng et~al.(2026)]{hiper}
Jiangweizhi Peng, Yuanxin Liu, Ruida Zhou, Charles Fleming,
Zhaoran Wang, Alfredo Garcia, and Mingyi Hong.
\newblock \emph{{HiPER}: Hierarchical Reinforcement Learning
with Explicit Credit Assignment for Large Language Model
Agents}.
\newblock In \emph{Proceedings of the 43rd International
Conference on Machine Learning (ICML)}, 2026.
\newblock \url{https://arxiv.org/abs/2602.16165}

\bibitem[Qi et~al.(2026)]{stapo}
Qiuyi Qi, Tian Liang, Mutian Bao, Jinjian Zhang,
Dongnan Liu, Wei Zhou, Linjian Mo, Ming Kong, Jie Liu,
Feng Zhang, and Qiang Zhu.
\newblock \emph{{STAPO}: Selective Trajectory-Aware Policy
Optimization for {LLM} Agent Training}.
\newblock In \emph{Proceedings of the 64th Annual Meeting of
the Association for Computational Linguistics (Volume 1:
Long Papers)}, pages 28371--28392. Association for
Computational Linguistics, 2026.
\newblock \url{https://aclanthology.org/2026.acl-long.1308/}

\bibitem[Wang et~al.(2026)]{beacon}
Zixuan Wang, Yuchen Yan, Hongxing Li, Teng Pan,
Dingming Li, Ruiqing Zhang, Weiming Lu, Jun Xiao,
Yueting Zhuang, and Yongliang Shen.
\newblock \emph{Milestone-Guided Policy Learning for
Long-Horizon Language Agents}.
\newblock In \emph{Proceedings of the 43rd International
Conference on Machine Learning (ICML)}, 2026.
\newblock \url{https://arxiv.org/abs/2605.06078}

\bibitem[Zhang et~al.(2025)]{rlvmr}
Zijing Zhang, Ziyang Chen, Mingxiao Li, Zhaopeng Tu, and
Xiaolong Li.
\newblock \emph{{RLVMR}: Reinforcement Learning with
Verifiable Meta-Reasoning Rewards for Robust Long-Horizon
Agents}.
\newblock arXiv:2507.22844, 2025.
\newblock \url{https://arxiv.org/abs/2507.22844}

\bibitem[Wang et~al.(2026a)]{envrl}
Zhitong Wang et al. 
\newblock \emph{{EnvRL}: Learn from Environment Dynamics in
Agentic Reinforcement Learning}.
\newblock arXiv:2606.17680, 2026.
\newblock \url{https://arxiv.org/abs/2606.17680}

\bibitem[Xie et~al.(2026)]{historytostate}
Haoyang Xie et al. 
\newblock \emph{From History to State: Constant-Context Skill
Learning for {LLM} Agents}.
\newblock arXiv:2605.05413, 2026.
\newblock \url{https://arxiv.org/abs/2605.05413}

\bibitem[Cho et~al.(2026)]{automata}
Seonglae Cho et al. 
\newblock \emph{Automata from Agent Traces: Failure and
Next-Step Prediction}.
\newblock arXiv:2608.23670, 2026.
\newblock \url{https://arxiv.org/abs/2608.23670}

\end{thebibliography}
\end{document}